%% file: 0_main.tex
\documentclass[letterpaper, 10 pt, conference]{ieeeconf}  

\IEEEoverridecommandlockouts                              

\usepackage{lipsum} 
\usepackage{bbold}
\usepackage{bm}
\usepackage{hyperref}
\usepackage{hhline} 
\usepackage{makecell} 
\usepackage{pifont}  
\usepackage[symbol]{footmisc}
\usepackage[table]{xcolor} 
\usepackage{colortbl}       
\usepackage{booktabs}       
\usepackage{pgf}
\definecolor{green1}{RGB}{0,100,0}   
\definecolor{green2}{RGB}{144,238,144} 

\usepackage[noadjust]{cite}

\usepackage{color,soul} 
\usepackage{indentfirst}

\usepackage{amsmath}

\title{\LARGE \bf  QDTraj: Exploration of Diverse Trajectory Primitives for Articulated Objects Robotic Manipulation }

\usepackage[dvipsnames]{xcolor} 

\newif\ifJoh
\newif\ifMat
\newif\ifFra
\newif\ifIgn

\Johtrue
\Mattrue
\Fratrue
\Igntrue

\begin{document}

\author{
Mathilde Kappel$^{1}$, 
Mahdi Khoramshahi$^{1}$, 
Louis Annabi$^{1}$, 
Faïz Ben Amar$^{1}$, 
Stéphane Doncieux$^{1}$\\[3pt]
$^{1}$ Institute of Intelligent Systems and Robotics, CNRS, Sorbonne University, Paris, F-75005, France\\ 
{\{kappel, khoramshahi, annabi, amar, doncieux\}@isir.upmc.fr }
}
\maketitle




\input{tex_files/abstract}

\input{tex_files/1_introduction}

\input{tex_files/2_related_works}

\input{tex_files/3_method}

\input{tex_files/4_experiments}

\input{tex_files/5_results_and_discussion}

\input{tex_files/6_conclusions}

\input{tex_files/acknowledgment}

\bibliographystyle{IEEEtran}

\input{tex_files/bilbio}
\end{document}

%% file: tex_files/abstract.tex

\begin{abstract}
Thanks to the latest advances in learning and robotics, domestic robots are beginning to enter homes, aiming to execute household chores autonomously. However, robots still struggle to perform autonomous manipulation tasks in open-ended environments. In this context, this paper presents a method that enables a robot to manipulate a wide spectrum of articulated objects. 

In this paper, we automatically generate different robot low-level trajectory primitives to manipulate given object articulations. A very important point when it comes to generating expert trajectories is to consider the diversity of solutions to achieve the same goal. Indeed, knowing diverse low-level primitives to accomplish the same task enables the robot to choose the optimal solution in its real-world environment, with live constraints and unexpected changes. To do so, we propose a method based on Quality-Diversity algorithms that leverages sparse reward exploration in order to generate a set of diverse and high-performing trajectory primitives for a given manipulation task. 

We validated our method, QDTraj, by generating diverse trajectories in simulation and deploying them in the real world. QDTraj generates at least 5 times more diverse trajectories for both hinge and slider activation tasks, outperforming the other methods we compared against. We assessed the generalization of our method over 30 articulations of the PartNetMobility articulated object dataset, with an average of 704 different trajectories by task. Code is publicly available at: \url{https://kappel.web.isir.upmc.fr/trajectory_primitive_website/}.

\end{abstract}

%% file: tex_files/1_introduction.tex

\section{INTRODUCTION}

Robotics applications have progressively evolved from controlled tasks toward increasingly unstructured tasks with higher levels of uncertainty. Accordingly, this paper focuses on robotic manipulation in everyday environments. Such environments contain a wide variety of articulated objects. This work, therefore, aims to address their manipulation.

Articulated objects contain one or more articulations, that is to say, one or more passive joints between two rigid parts of the same object entity~\cite{Liu2025ArticulatedObjects}. For example, an oven is an articulated object with a hinge joint connecting the frame to the door and a slider joint allowing the oven rack to move in and out.

One of the main challenges of contact-rich articulated object manipulation relies on the expert data acquisition challenge. Current approaches intend to acquire reliable expert manipulation demonstrations through a teleoperated data acquisition system~\cite{OpenXEmbodiment2025,LuoHumanInTheLoop}, or simulated data collection~\cite{GaoCompositionalGeneralization2024}. However, real-world data collection is time-consuming and expensive, which motivates the use of simulated data. Simulated data, however, can suffer from sim-to-real inconsistencies. To address this, our approach leverages compliant control to absorb modeling errors until what compliance can handle; beyond that, another trajectory can be easily chosen from all the diverse solutions we generated.  The objective of this paper is to address these limitations by automatically generating multiple physically grounded expert demonstrations in simulation. In this work, we generate low-level manipulation demonstrations that combine high-quality initial grasp synthesis with reliable contact-rich interactions throughout the entire manipulation motion.

\input{tex_files/figures/APGenerator_Intro}

Another major challenge arising from the difficulty of data acquisition that we address is ensuring diversity among demonstrations aquisition. Many works propose methodologies for manipulation~\cite{Kroemer2020RobotLearning}, but emphasizing the importance of redundancies and novelty~\cite{Oudeyer2009IntrinsicMotivation,Lynch2019LatentPlans} remains underexplored for specific articulated object manipulation. Redundancies refer to the fact that a single manipulation task can be achieved in multiple different ways. Considering these redundancies enables the deployment of manipulation tasks in open-ended scenarios, as different trajectories may be more appropriate depending on the context, situation, higher-level task, or uncertainty. To achieve this, we leverage Quality-Diversity (QD) algorithms, a family of evolutionary methods known for their efficient exploration capabilities~\cite{Cully2017QDFramework}. As such, QD, a gradient-free learning strategy, provides a powerful framework to discover multiple ways of manipulating the same object.

A third challenge addressed in this paper is the inherent complexity of tasks involving articulated objects, arising from the multiplicity of uses and contexts in which such manipulation occurs. This work tackles this difficulty by breaking down a complex manipulation task into different tasks with a single degree of freedom activated at a time. For example, when removing the oven rack, a robot must first open the oven door by applying a force on the handle that moves the hinge joint between the door and the frame, which constitutes the first task. It then pulls out the oven rack by activating the slider joint as the second task.

By addressing the three challenges outlined above, we propose a plug-and-play methodology to automatically generate diverse low-level primitives for articulated object manipulation (cf Figure~\ref{fig:APEvo}). Central to our approach is the first application of a Quality-Diversity (QD) algorithm to trajectory generation primitives for articulated objects. Our algorithm, QDTraj (Quality-Diversity-Trajectory, produces a set of high-performing, diverse adaptive robot trajectories capable of manipulating articulated objects. 

To illustrate the efficiency of our method, we generate primitives in simulation, perform a comparative study, and deploy the resulting trajectories on a real robot for our experimental objects.

%% file: tex_files/figures/APGenerator_Intro.tex
\begin{figure}[t]
  \centering
  \includegraphics[width=\columnwidth]{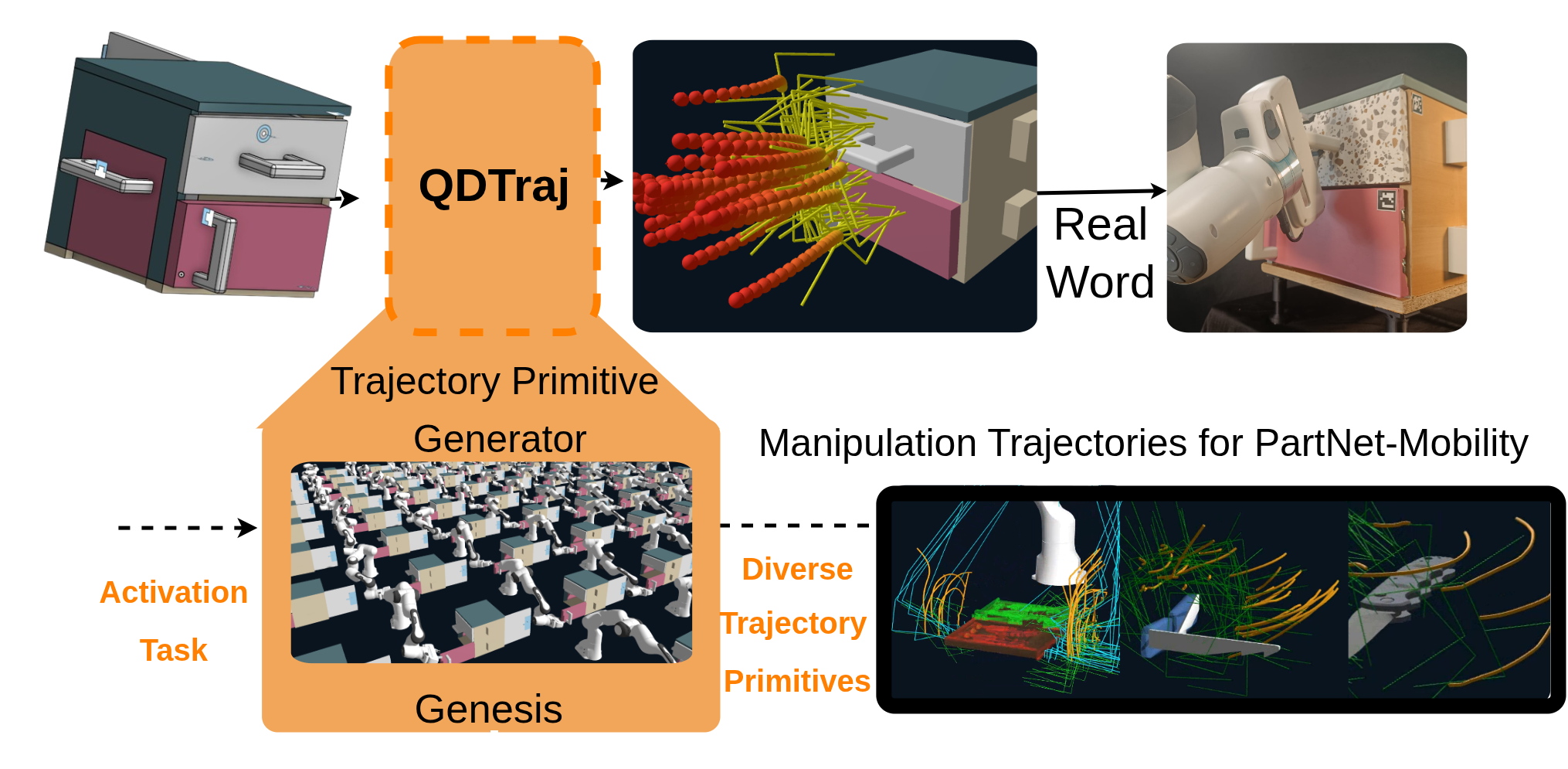}
  \caption{\textbf{Plug-and-play QDTraj exploration algorithm.} Given an articulated object URDF and an activation task, QDtraj generates sets of diverse trajectory primitives to achieve the task. Trajectory primitives generated in Genesis parallelized simulation are deployable real world set-up.}
  \label{fig:APGenerator_intro_lab}
\end{figure}

%% file: tex_files/2_related_works.tex

\section{RELATED WORKS}
\label{sec:2_related_works}

\textbf{\textit{Articulated objects Dataset.}} Datasets of 3D articulated objects are commonly used to facilitate learning-based approaches for manipulation tasks involving articulated structure \cite{liu2024survey}. In our work, we use PartNet-Mobility~\cite{xiang2020sapien}, a dataset of 2,346 articulated objects across 46 categories and designed for robotics manipulation research and simulation in environments. Better than PartNet \cite{mo2019partnet}, which only annotated indoor object parts, PartNet-Mobility provides 3D models of articulated objects along with annotations for movable parts and their joint types (hinge, slider, screw). 

\textbf{\textit{Articulated Objects Learning.}}
There are learning methods for object manipulation that rely on different interaction policies to discover actionable grasp areas. As an example, AO-Grasp~\cite{morlans2025aograsp} provides both stable and actionable grasps. Similarly, GAPartNet~\cite{geng2023gapartnet} provides actionable grasps and interaction policies that enable cross-category object manipulation. However, its policies are highly hand-designed and geometry-specific (\textit{Round Fixed Handle}, \textit{Line Fixed Handle}, \textit{Hinge Handle}, \textit{Slider Button}, etc.), which limits adaptability to variations such as misaligned handles, partially blocked objects, or missing handles. Another example is PartManip~\cite{geng2023partmanip} use Proximal Policy Optimization (PPO)~\cite{PPO2017}, with rotational and distance-based rewards to learn the interaction policy. Lastly, AdaRPG~\cite{zhang2025adamanip} leverages foundation models to learn grasp areas by defining six atomic manipulation functions in the end-effector local frame; this approach still restricts the robot to predefined orientations and action types. Another work, Where2Act~\cite{mo2021where2act}, explicitly mentions that half its offline data relies on randomly sampling action primitives, resulting in low expert sampling efficiency (1\%), while the other half uses subregion-adaptive sampling that is not explicitly diversity-focused. Where2Explore~\cite{ning2023where2explore} also relies on six types of hard-coded Cartesian motion primitives. In contrast, VAT-Mart~\cite{wu2022vatmart} predicts residual gripper poses along task trajectories, producing more realistic motions, yet the diversity in this approach only affects trajectories rather than initiation points for interaction. 
Our method identifies interactive regions using an efficient strategy that encourages diversity and does not rely on precomputed interaction trajectories, enabling fully adaptive manipulation of articulated objects.

\textbf{\textit{Compliant Policies}}
Adaptive control strategies enable robots to interact compliantly with dynamic and uncertain physical environments, as demonstrated in works such as \cite{zeng2021learning}. Our contribution also uses compliant control for the same reason. AdaManip~\cite{wang2025adamanip} focuses on learning adaptive manipulation policies after having identified five adaptive mechanisms (Lock Mechanism, Random Rotation Direction, Rotate  Slide Mechanism, Push/Rotate Mechanism, Switch Contact Mechanism).

\textbf{\textit{Leveraging learning with Diversity}}
\input{tex_files/figures/primitive}
The lack of diversity has been identified as a major limitation in the learning of articulated object manipulation. Quite recently works~\cite{shi2025diversity}~\cite{havrilla2024synthetic} highlighted that diversity plays a crucial role in robotic learning. Similarly,  Y. Hu et al. \cite{Hu2025DataScaling} demonstrated the link between the generalization capability of robot policies and the number of environments, objects, and demonstrations, offering principles for efficient data collection strategies. Moreover, M. Shi et al.~\cite{shi2025diversity} showed that task diversity directly benefits robotic performance and generalization. In that respect, some work focuses on intrinsically motivated goal exploration mechanisms~\cite{baranes2010intrinsically}. Closely related, the family of Quality-Diversity (QD) optimization algorithms differs from traditional methods that search for a single optimal solution, as QD algorithms aim to discover a diverse repertoire of high-performing solutions~\cite{Cully2017QDFramework}. Among them, MAP-Elites~\cite{mouret2015mapelites} is a grid-based QD algorithm that systematically explores the solution space. QDG6-DoF \cite{huber2024speeding} extends Quality-Diversity optimization to robotic grasping. QDGSet \cite{huber2024qdgset} further extends it by providing a framework for data augmentation, but focuses on grasping and  not articulated object manipulation trajectories. R. Zurbrügg et al.~\cite{zurbrugg2026dexevolve} uses an evolutionary algorithm for complex dexterous grasping synthesis. However, these approaches do not address trajectory generation after grasp synthesis. In contrast, QDTraj algorithm goes beyond grasping to generate full manipulation trajectories. \\

%% file: tex_files/figures/primitive.tex
\begin{figure}[t]
  \centering
  \includegraphics[width=\columnwidth]{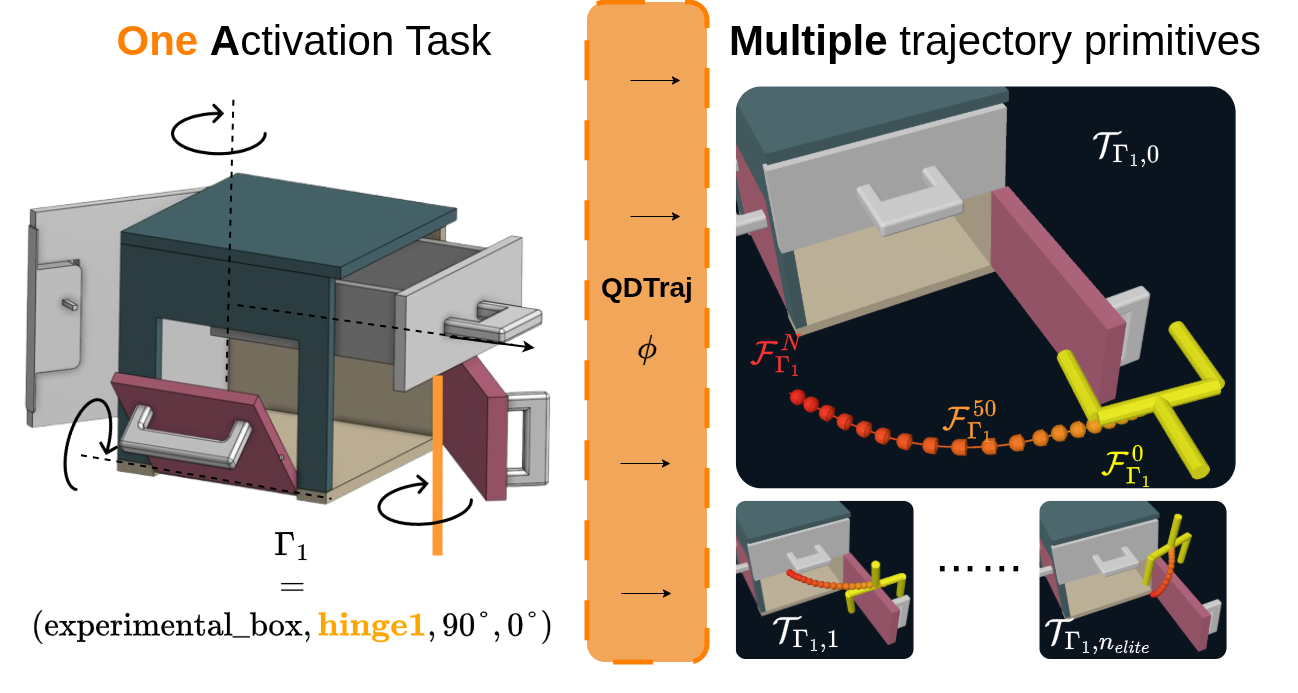}
  \caption{\textbf{Input/Output of QDtrajAction plug-and-play module}
 As input the Experimental Box URDF and an activation task to accomplish. In orange on the CAD is the axis of revolution of the hinge involved in the task input $\Gamma_1$, the other joints in black are ignored. As the output a set of trajectory primitives, each trajectory has a grasp starting frame (in yellow) as well as a full trajectory (in gradient orange dot).}
  \label{fig:primitive}
\end{figure}

%% file: tex_files/3_method.tex

\section{METHOD}
\label{sec:3_method}

QDTraj enables the exploration of diverse trajectory primitives to execute a given activation task on an articulated object. We make the assumption of having URDF of articulated objects provided upstream by prior knowledge or by an upstream reconstruction module~\cite{gupta2026pokenet}. First, Section ~\ref{subsec:Articulation_Aware_Formulation} describes our formulation regarding the representation of an activation task and trajectory primitives. Then Section ~\ref{subsec:QualityDiversity_Formulation} describes our Quality-Diversity formulation for QDTraj. Lastly, Section ~\ref{subsec:Contribution_adaptative_trajectory} provides more details about our QDTraj evaluation function, which is our own interaction strategy.  

\subsection{Movement primitive representation}
\label{subsec:Articulation_Aware_Formulation}

\textbf{\textit{Action primitive:}} In this work, we adopt an articulation-centered oriented classification for task primitives. Each robot activation task is defined with respect to the object articulation it enables to activate on a given object. We consider single–degree-of-freedom object joints, such as prismatic or revolute joints. Building upon this articulation-centered classification, we explore prehensile manipulation, involving contact-rich manipulation between the articulated objects and the robot end-effector. 

Each activation task is parameterized by four parameters: the articulated object to be manipulated, the joint the robot has to interact with, and the initial and target values of this articulation configuration in the scene. Formally, we define an activation task as follows:
\begin{equation}
\Gamma = (\text{object}, \text{articulation+index}, s_{\text{init}}, s_{\text{target}}) \in \mathcal{X} \times \mathcal{A} \times \mathbb{R} \times \mathbb{R}
\end{equation}
with $\mathcal{X}$ a set of articulated objects and $\mathcal{A}$ a set of two articulation types: prismatic or revolute.

\textbf{\textit{Trajectory primitive:}} QDTraj generates a set of diverse trajectory primitives able to achieve the same activation task. Each trajectory primitive corresponds to the path of the robot end effector that enables the realisation of a given task on the object to manipulate.
That is to say, each trajectory primitive $\mathcal{T}_\Gamma$ is formalized as a sequence of robot end-effector positions in the object reference frame: 
\begin{equation}
\mathcal{T}_\Gamma = \left\{ \mathcal{F}^{(k)}_{ee/\text{object}} \right\}_{k=0}^{N}
\end{equation}
\noindent
where $\mathcal{F}_{\text{eef}/\text{object}}$ denotes the pose of the end effector expressed in the object frame at step $k$ over $N$ total trajectory steps.

QDTraj algorithm can be formalized as a function $\phi$ that takes as input an activation task $\Gamma$ and returns as output a set of diverse trajectory primitives to accomplish this activation task.

QDTraj algorithm enables the exploration of $n_{\text{elite}}$ alternative trajectories to achieve the same task: $n_{\text{elite}} = \mathrm{Card}\bigl(\phi(\Gamma)\bigr)$.
Figure \ref{fig:primitive} illustrates input/output QDTraj. 

\subsection{Learning algorithm}
\label{subsec:QualityDiversity_Formulation}

\input{tex_files/figures/APEvo}

The diversity of solutions generated by QDTraj comes from the use of a QD algorithm called MAP-Elites~\cite{mouret2015mapelites}. The output of QDTraj is the structured archive of high-scoring solutions built with MAP-Elites. Our contribution lies in the design of the genotype, mutation operator, evaluation function, selection method, and behavioral descriptors used in the MAP-Elites algorithm. These design choices are explained in this section.

For a given robot arm, QDTraj takes as input an activation task $\Gamma$, and returns a set of trajectory primitives for the robotic arm to accomplish the action primitive $ \phi(\Gamma)$. At initialization, a set of $N$ end effector positions is initialized around the target object. Each of these 6-DoF frames defines a grasp candidate and serves as the starting key-frame of a trajectory. Formally, each trajectory starting frame candidate is an individual $i$ parameterized by the following genotype:
\begin{equation}
    g_i = (x_i,y_i,z_i,q_{1i},q_{2i},q_{3i},q_{4i}) \in \mathbb{R}^3 \times [-1,1]^4
\end{equation}
\noindent

Where $(x_i,y_i,z_i)$ represents the cartesian position of the end effector frame in the object reference frame and $(q_{1i},q_{2i},q_{3i},q_{4i})$ the orientation parameters unit quaternion. 

Then, QDTraj relies on the selection-mutation loop (illustrated Fig. ~\ref{fig:APEvo}), with the key steps described below: 

\begin{enumerate}    
    \item Parallelized Evaluation.
    Each end effector starting grasp candidate is evaluated in parallel using $N$ simulation environments. For each individual, a relative trajectory $ \mathcal{T}_i$ is generated through our adaptive evaluation process (Section ~\ref{subsec:Contribution_adaptative_trajectory}). After that, a fitness score $f_i$ is affected. Figure ~\ref{fig:metrics} illustrates our fitness assessment for each trajectory. 
    
    \item MAP-Elites Behavioral Descriptors.
   Individuals are stored in a MAP-Elites archive according to their behavioral descriptor and fitness value (defined in Figure~\ref {fig:metrics}). The behavioral descriptor of an individual is defined as the three-dimensional Cartesian starting position of its genotype, $\mathbf{b}(g_i) = \mathbf{b}_{xyz}$, rounded to the nearest hundredth of a centimeter. We can represent the output archive in 3D space as a grid $\mathcal{B}$ composed of small cubes of 1cm side. If two individuals have the same behavioral descriptor, a local competition is performed within their 3D cell, and only the individual with the highest fitness score is kept in the archive. 

    \item Selection.
    In our contribution, we fill the new population of individuals by selecting half of the genotypes from the highest performing MAP-Elites cells, and half randomly. Variants of this selection process have also been compared in the experimental ablation section ~\ref{subsec:ablation})

    \item Mutation.
    Each selected individual from the new selected population is individually mutated to form offspring, creating the final new population of candidate starting frame trajectories. Concretely, we add a bounded noise on the Cartesian coordinates as well as on the quaternion orientation of starting poses.     
\end{enumerate}
The algorithm iterates between mutation, evaluation, and selection phases for $n_{\text{generations}}$ generations, progressively refining the archive $A$.
At the end of the evolutionary process, QDTraj produces a MAP-Elites archive $A$ containing $n_{\text{elite}}$ diverse and high-performing trajectory primitives. The output archive $\mathcal{A}$ can be formalized as follows:

\begin{equation}
\forall \mathbf{b}_{xyz} \in \mathcal{B}, \quad
\mathcal{A}_{{b}_{xyz}} =
\begin{cases}
(g_{i_{\mathrm{elite}}}, \mathcal{T}_{i_{\mathrm{elite}}}, f_{i_{\mathrm{elite}}}),\\
i_{\mathrm{elite}} = \displaystyle \arg\max_{i \,:\, \mathbf{b}(g_i) = \mathbf{b}_{xyz}} f_i,
\end{cases}
\end{equation}

\noindent
where: $g_{i_{\mathrm{elite}}}$ is the grasp starting point of the genome maximizing the fitness in cell $\mathbf{b}_{xyz}$, $\mathcal{T}_{i_{\mathrm{elite}}}$ is the full trajectory associated with this genome, and  $f_{i_{\mathrm{elite}}}$ is the fitness score of this genome.

\subsection{Adaptive Interaction policy}
\label{subsec:Contribution_adaptative_trajectory}
\input{tex_files/figures/metrix}
QDTraj evaluation function relies on an adaptive interaction policy. Given a joint activation task $\Gamma$ and a candidate starting point $g_i$, QDTraj evaluation function generates an adaptive prehensile trajectory $\mathcal{T}_i$.

During our adaptive interaction strategy, for each trajectory starting point, the corresponding robot configuration is computed via inverse kinematics. Once this joint configuration is reached, the two-fingers end effector is closed using force closure.  

If the utility surfaces of the gripper (in yellow Figure ~\ref{fig:metrics}) are in contact with the utility surfaces of the object part to be grasped (part mesh), the grasp is considered successful by QDTraj contact evaluation function. After that, the robot is switched to compliant mode: the gains of the robot joints are set very low. In compliant control mode, the robot responds softly to external forces on its joints, as the low robot gains reduce stiffness and damping. Meanwhile, the gains of the object joint of interest are set very high, and this joint is actuated from $s_{\text{init}}$ to $s_{\text{target}}$ through intermediate joint target values. High object gains ensure precise control of the object parts' motion.

During this virtual object joint actuation, QDTraj contact function continuously monitors the interaction between the end effector. At the end of the object actuation, the complete end effector trajectory resulting from the guiding of the object over the robot is recorded, along with the percentage $f_i$ of the articulation value at which detachment occurred between the robot and the object.

%% file: tex_files/figures/APEvo.tex
\begin{figure}[t]
  \centering
\centering
  \includegraphics[width=\columnwidth]{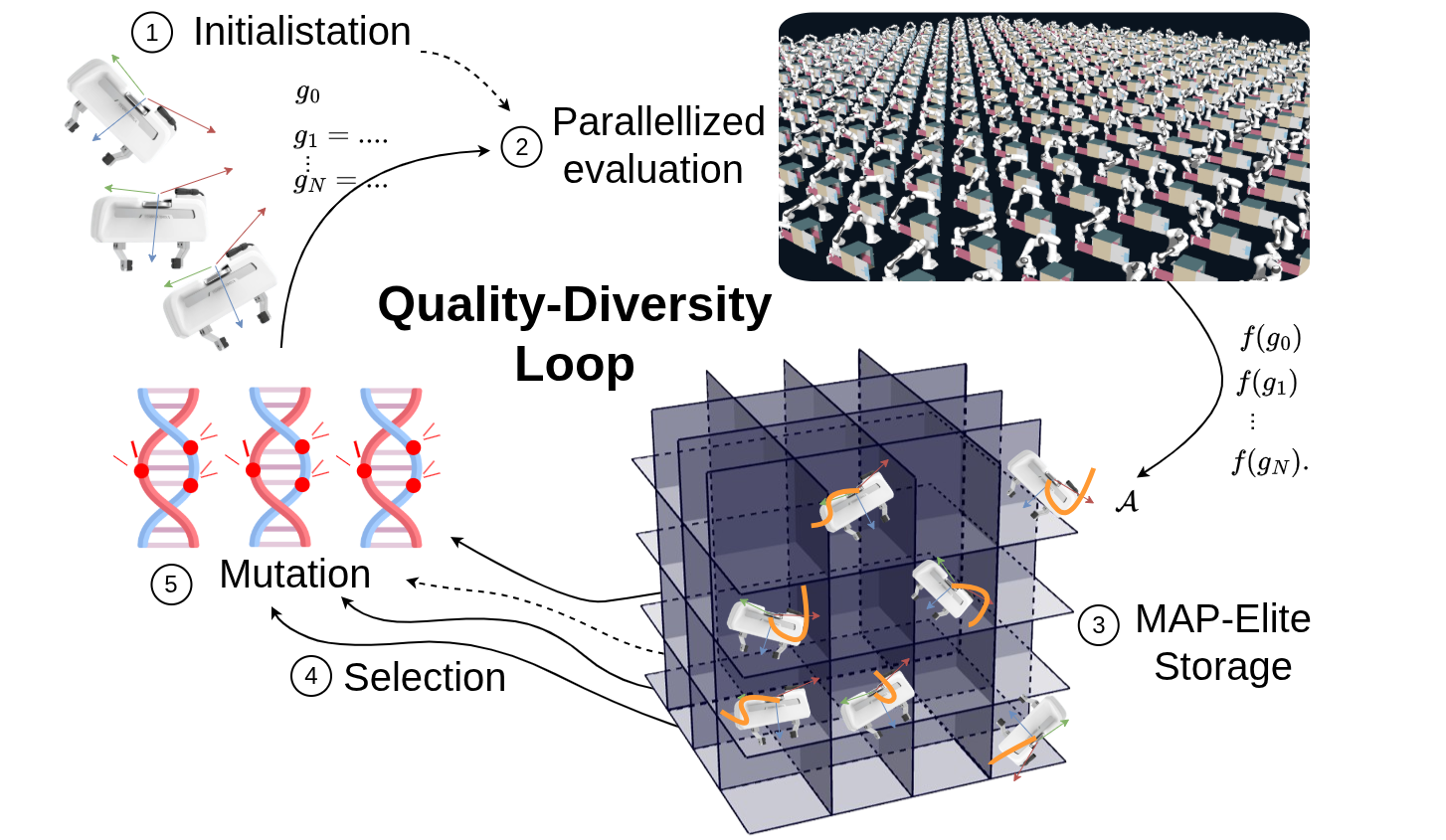}
  \caption{\textbf{QDtraj selection-mutation loop} \textbf{(1)} Each genotype is a FR3 end effector frame \textbf{(2)} Parallelized evaluation in 10 000 Genesis simulated environments, \textbf{(3)} MAP-Elites 3D archive $\mathcal{A}$, each small 3D cell in the 3D grid represents a different behavior descriptor $b_ {xyz}$ \textbf{(4)} Selection of individuals \textbf{(5)} Mutation of the genotypes }
  \label{fig:APEvo}
\end{figure}

%% file: tex_files/figures/metrix.tex
\begin{figure}[t]
  \centering
\centering
  \includegraphics[width=\columnwidth]{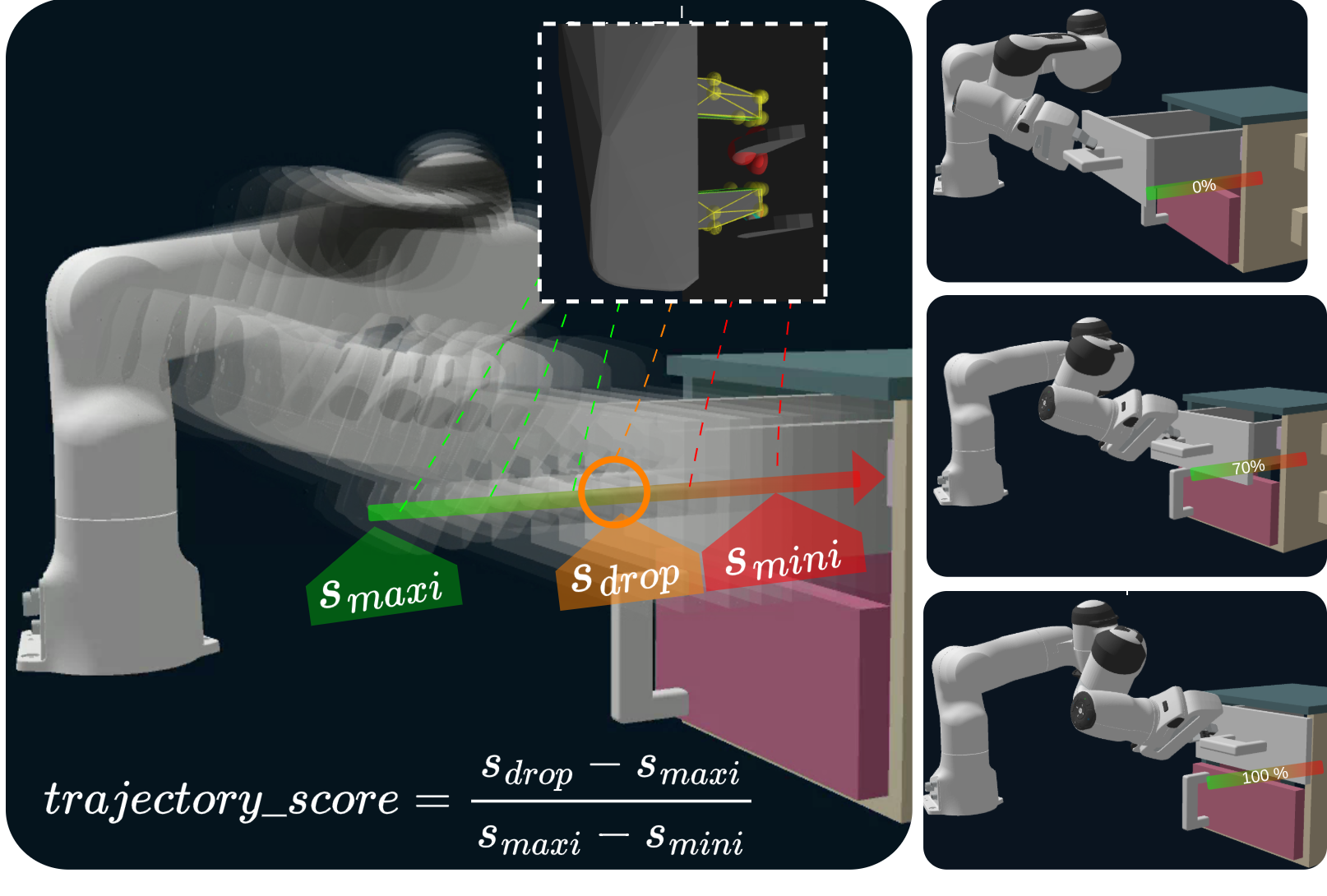}
  \caption{\textbf{Fitness function to assert the quality of each trajectory explored}. The score of one trajectory is calculated regarding the joint value of the articulation at the end of the trajectory. It is the ratio between the angular displacement traveled $s_{drop}-s_{maxi}$ and the total angular displacement to be traveled $s_{maxi} -s_{mini}$.}   
  \label{fig:metrics}
\end{figure}

%% file: tex_files/4_experiments.tex

\section{EXPERIMENTS}
\input{tex_files/figures/visual_baseline_traj}

\input{tex_files/figures/results}

\subsection{Experimental set-ups}
For both simulation and real-world experiments, we used a Franka Emika FR3 robotic arm equipped with a two-fingers gripper. 
Since there is no standard articulated object dataset in the real world, we tested our method on a custom experimental object.  We created a CAD model faithful to the real-world properties of our experimentally designed articulated object, and an URDF model was extracted thanks to onshape-to-robot~\cite{onshape-to-robot}.  

Besides, in order to demonstrate the generalization of our plug-and-play QDTraj algorithm, we also ran simulations on other objects from an off-the-shelf articulated object dataset: PartNet-Mobility~\cite{xiang2020sapien}. 
All trajectories were generated using the Genesis physics engine. We implemented our algorithm using~\cite{Genesis} in 10,000 parallelized environments on GPU. 

\subsection{Baselines and ablations }
\label{subsec:ablation}
\input{tex_files/tables/table_quantitative}
Most works related to our contribution, reviewed in Section~\ref{sec:2_related_works}, focus on either a learning-based grasp prediction backbone and/or a trajectory interaction policy. To enable meaningful and fair comparisons, we lead an experimental protocol structured along the two orthogonal axes: grasp strategy and trajectory strategy. 

\textbf{\textit{Interaction Policy Baselines}} (QDTraj Evaluation process).
On the first axis, we evaluate our proposed adaptive interaction strategy against other literature interaction policies' action space. Formally, this corresponds to modifying the trajectory QDTraj evaluation step while keeping all the other stages of QDTraj formulation.

Instead of using QDTraj’s native interaction strategy, we inject trajectories generated by baseline interaction policies' action spaces. 

The following three interaction policy action spaces (illustrated Figure~\ref{fig:visual_baseline_traj}) are implemented: :
\begin{itemize}
    \item Adaptive-AS (Ours): Adaptive Interaction policy Action Space. Newly introduced in this paper, detailed in Section~\ref{subsec:Contribution_adaptative_trajectory}.
    \item Where2Act-AS: Where2Act Interaction policy Action Space.
    The six predefined action primitives described in Where2Act~\cite{mo2021where2act}. These primitives define the interaction action space after grasping. Each primitive is parameterized in $\mathrm{SE}(3)$ and executed from the grasp starting pose using hard-coded motion trajectories. The six primitives correspond to fixed end-effector directions:
    $\text{Push } (1,0,0)$, $\text{Pull } (-1,0,0)$, $\text{Right } (0,1,0)$, $\text{Left } (0,-1,0)$, $\text{Up } (0,0,1)$, $\text{Down } (0,0,-1) $

    \item VAT-Mart-AS: Interaction Policy Action Space.
    VAT-Mart~\cite{wu2022vatmart} trajectory action space, where each interaction is defined as a sequence of 6-DoF end-effector waypoints with variable length. Each waypoint consists of a 3-DoF position and a 3-DoF orientation, represented using the 6D rotation representation. Subsequent waypoints are generated by randomly sampling residual gripper poses, resulting in stochastic step-by-step trajectories.
\end{itemize}

\textbf{\textit{Grasp Strategy Ablation}} (QDTraj Selection process).
On the second axis, we evaluate the impact of the grasp strategy through an ablation study within the QDTraj framework. Formally, this study corresponds to changing the selection mechanism (step 4) used during quality-diversity optimization, while keeping all the other QDTraj steps fixed.

The following three grasp strategy variants are implemented:
\begin{itemize}
    \item MAP-Elites Explore (Ours): Detailed in ~\ref{sec:3_method}, step (3).
    \item MAP-Elites Success:
    Similar to MAP-Elites Explore, but without cloning successful individuals.
    \item Random Search: 
    Individuals are randomly sampled at each generation, without selection or mutation. This variant serves as a lower-bound baseline.
\end{itemize}

Hereafter, we refer to a QDTraj variant algorithm configuration as a specific combination of a grasp selection strategy and an interaction strategy. In the comparative study we conduct, we systematically evaluate all possible combinations of the three grasp strategies and the three interaction strategies described above. 
\subsection{Evaluation protocol}

To evaluate our approach, we conducted experiments following two complementary protocols. The first protocol focuses on a controlled and detailed comparison of all nine variant algorithm configurations on our experimental object. In contrast, the second protocol evaluates the generalization capability of our method across a wider range of PartNetMobility articulated objects for our contribution algorithm configuration QDTraj-MAP-Elites-Explore-Adaptive-AS.

In the first protocol, two tasks are considered: $\Gamma_1$, the activation of the front hinge on the experimental box from $90^\circ$ to $0^\circ$, and $\Gamma_2$, the activation of the upper slider on the experimental box from 0\,cm to 20\,cm. Each run is executed for $175$ generations. To improve statistical significance, during all experimental phases, each run is repeated using ten independent random seeds, and the reported results correspond to the average over these ten seeds. Overall, this first experimental protocol involves a total of $
3 \text{ grasp methods} \times 3 \text{ trajectory methods} \times 2 \text{ tasks} \times 10 \text{ seeds} = 180 \text{ runs.} $

In the second protocol, which focuses on our proposed algorithm configuration, each run is executed for 50 generations. Overall, we evaluate 30 action primitives across 10 different objects from the PartNet-Mobility dataset, resulting in the generation of more than 10,000 trajectory primitives. The evaluated articulated objects are:
two toasters, one microwave, three faucets,
two dispensers, one kitchen pot, and one coffee machine.

\subsection{Evaluation metrics } 
\textbf{\textit{Trajectory metrics:}}. We evaluate each run by assigning the number of successful trajectories $\mathcal{T}$ explored in its outcome archive $A$. One trajectory success is considered if the trajectory primitive allows for achieving 65 percent of the joint value activation requirement ($f_{i_{\mathrm{elite}}}>=65\%$). Figure \ref{fig:metrics} illustrates the trajectory metrics. 
Our framework provides a replay mode in which the robot executes the recorded end-effector trajectory while the articulated object remains passive, enabling visualization of each successful object-guided trajectory in simulation as a robot-guided object motion.

\textbf{\textit{Grasp metrics: }} As complementary information, we count the number of successful grasps, found in the output archive $\mathcal{A}$, even if they do not lead to a successful trajectory. 

Owing to the discretization of the behavioral space $\mathcal{B}$, each occupied cell of the MAP-Elites archive $\mathcal{A}$ stores a single elite primitive selected by fitness maximization. Consequently, the number of successful grasps stored in the archive directly reflects the spatial diversity of the generated trajectory primitives.

\textbf{Real Robot}:
To ensure sim-to-real transfer, all movements are deployed with ROS2 Cartesian controller for Franka FR3 with impedance control.

%% file: tex_files/figures/visual_baseline_traj.tex
\begin{figure}[t]
  \centering
\centering
  \includegraphics[width=\columnwidth]{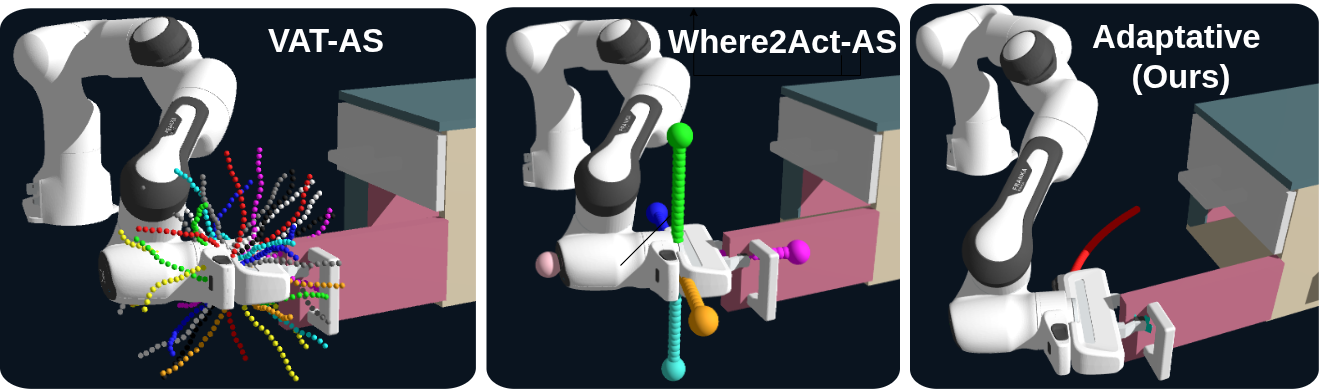}
  \caption{\textbf{Baseline for interaction action spaces} \textbf{Left.} VAT-Mart interaction policy Action Space. \textbf{Middle.} Where2Act interaction policy Action Space, \textbf{Right.} The Adaptive interaction policy Action Space introduced in this work. }
  \label{fig:visual_baseline_traj}
\end{figure}

%% file: tex_files/figures/results.tex
\begin{figure*}[t]
  \centering
\centering
  \includegraphics[width=\textwidth]{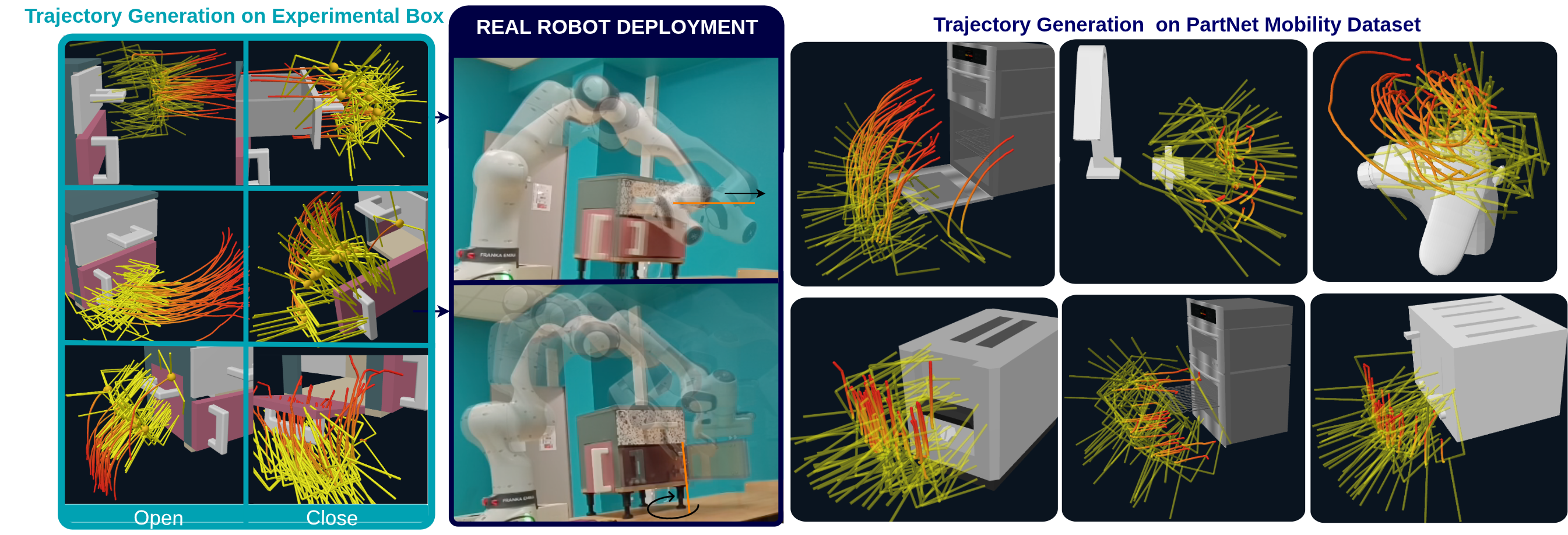}
  \caption{\textbf{Simulation and real world Experiments - Qualitative results} \textbf{Left.} Six output archives $\mathcal{A}$ of six QDtraj runs concerning six different task primitives related to our Experimental Box. \textbf{Middle.} Real world deployement of two original trajectories generated by QDtraj. \textbf{Right.} Output archives $\mathcal{A}$ of 6 QDtraj runs concerning task primitives related to PartNetMobility Objects }
  \label{fig:results}
\end{figure*}

%% file: tex_files/tables/table_quantitative.tex
\begin{table*}[t]
\centering
\begin{tabular}{lcccccccc}
\toprule
\textbf{Method} 
& \multicolumn{2}{c}{\textbf{Adaptive-AS(Ours)}} 
& \multicolumn{2}{c}{\textbf{Where2Act-AS }} 
& \multicolumn{2}{c}{\textbf{VAT-Mart-AS }}
& \multicolumn{2}{c}{\textbf{Row Mean}} \\
\cmidrule(lr){2-3} \cmidrule(lr){4-5} \cmidrule(lr){6-7} \cmidrule(lr){8-9}
& \textbf{nbr grasps} & \textbf{nbr traj}
& \textbf{nbr grasps} & \textbf{nbr traj}
& \textbf{nbr grasps} & \textbf{nbr traj} 
& \textbf{nbr grasps} & \textbf{nbr traj} \\
\midrule

\multicolumn{9}{c}{\textbf{Hinge0 ($\Gamma_1$)}} \\

MAP-Elite-Explore  
& \cellcolor{blue!50!white}$3515\,\mathord{\scriptscriptstyle\textcolor{black!80}{\pm 202}}$
& \cellcolor{green!50!white}$\textbf{2944}\,\mathord{\scriptscriptstyle\textcolor{black!80}{\pm 193}}$
& \cellcolor{blue!50!white}$3550\,\mathord{\scriptscriptstyle\textcolor{black!80}{\pm 169}}$
& \cellcolor{green!20!white}$519\,\mathord{\scriptscriptstyle\textcolor{black!80}{\pm 92}}$
& \cellcolor{blue!45!white}$3343\,\mathord{\scriptscriptstyle\textcolor{black!80}{\pm 258}}$
& \cellcolor{green!15!white}$311\,\mathord{\scriptscriptstyle\textcolor{black!80}{\pm 60}}$
& $3470\,\mathord{\scriptscriptstyle\textcolor{black!60}{\pm 210}}$ 
& $\textbf{1258}\,\mathord{\scriptscriptstyle\textcolor{black!60}{\pm 92}}$ \\

MAP-Elite-Success  
& \cellcolor{blue!15!white}$895\,\mathord{\scriptscriptstyle\textcolor{black!80}{\pm 160}}$
& \cellcolor{green!25!white}$657\,\mathord{\scriptscriptstyle\textcolor{black!80}{\pm 124}}$
& \cellcolor{blue!20!white}$1152\,\mathord{\scriptscriptstyle\textcolor{black!80}{\pm 213}}$
& \cellcolor{green!5!white}$27\,\mathord{\scriptscriptstyle\textcolor{black!80}{\pm 10}}$
& \cellcolor{blue!18!white}$1126\,\mathord{\scriptscriptstyle\textcolor{black!80}{\pm 150}}$
& \cellcolor{green!3!white}$14\,\mathord{\scriptscriptstyle\textcolor{black!80}{\pm 5}}$
& $1058\,\mathord{\scriptscriptstyle\textcolor{black!60}{\pm 174}}$ 
& $233\,\mathord{\scriptscriptstyle\textcolor{black!60}{\pm 49}}$ \\

Random  
& \cellcolor{blue!5!white}$116\,\mathord{\scriptscriptstyle\textcolor{black!80}{\pm 15}}$
& \cellcolor{green!15!white}$66\,\mathord{\scriptscriptstyle\textcolor{black!80}{\pm 12}}$
& \cellcolor{blue!8!white}$232\,\mathord{\scriptscriptstyle\textcolor{black!80}{\pm 17}}$
& \cellcolor{green!2!white}$3\,\mathord{\scriptscriptstyle\textcolor{black!80}{\pm 1}}$
& \cellcolor{blue!8!white}$244\,\mathord{\scriptscriptstyle\textcolor{black!80}{\pm 27}}$
& \cellcolor{green!2!white}$3\,\mathord{\scriptscriptstyle\textcolor{black!80}{\pm 1}}$
& $197\,\mathord{\scriptscriptstyle\textcolor{black!60}{\pm 19}}$ 
& $24\,\mathord{\scriptscriptstyle\textcolor{black!60}{\pm 11}}$ \\

\midrule
\multicolumn{9}{c}{\textbf{Slider0 ($\Gamma_2$)}} \\

MAP-Elite-Explore  
& \cellcolor{blue!60!white}$5784\,\mathord{\scriptscriptstyle\textcolor{black!80}{\pm 461}}$
& \cellcolor{green!60!white}$\textbf{3988}\,\mathord{\scriptscriptstyle\textcolor{black!80}{\pm 296}}$
& \cellcolor{blue!55!white}$6652\,\mathord{\scriptscriptstyle\textcolor{black!80}{\pm 222}}$
& \cellcolor{green!50!white}$2489\,\mathord{\scriptscriptstyle\textcolor{black!80}{\pm 223}}$
& \cellcolor{blue!58!white}$6250\,\mathord{\scriptscriptstyle\textcolor{black!80}{\pm 259}}$
& \cellcolor{green!55!white}$2758\,\mathord{\scriptscriptstyle\textcolor{black!80}{\pm 345}}$
& $6229\,\mathord{\scriptscriptstyle\textcolor{black!60}{\pm 314}}$ 
& $\textbf{3078}\,\mathord{\scriptscriptstyle\textcolor{black!60}{\pm 212}}$ \\

MAP-Elite-Success  
& \cellcolor{blue!20!white}$906\,\mathord{\scriptscriptstyle\textcolor{black!80}{\pm 137}}$
& \cellcolor{green!30!white}$435\,\mathord{\scriptscriptstyle\textcolor{black!80}{\pm 80}}$
& \cellcolor{blue!35!white}$2818\,\mathord{\scriptscriptstyle\textcolor{black!80}{\pm 441}}$
& \cellcolor{green!30!white}$500\,\mathord{\scriptscriptstyle\textcolor{black!80}{\pm 121}}$
& \cellcolor{blue!32!white}$2588\,\mathord{\scriptscriptstyle\textcolor{black!80}{\pm 379}}$
& \cellcolor{green!30!white}$532\,\mathord{\scriptscriptstyle\textcolor{black!80}{\pm 121}}$
& $2104\,\mathord{\scriptscriptstyle\textcolor{black!60}{\pm 319}}$ 
& $489\,\mathord{\scriptscriptstyle\textcolor{black!60}{\pm 100}}$ \\

Random  
& \cellcolor{blue!10!white}$404\,\mathord{\scriptscriptstyle\textcolor{black!80}{\pm 19}}$
& \cellcolor{green!18!white}$186\,\mathord{\scriptscriptstyle\textcolor{black!80}{\pm 14}}$
& \cellcolor{blue!12!white}$667\,\mathord{\scriptscriptstyle\textcolor{black!80}{\pm 27}}$
& \cellcolor{green!12!white}$65\,\mathord{\scriptscriptstyle\textcolor{black!80}{\pm 11}}$
& \cellcolor{blue!12!white}$654\,\mathord{\scriptscriptstyle\textcolor{black!80}{\pm 36}}$
& \cellcolor{green!12!white}$84\,\mathord{\scriptscriptstyle\textcolor{black!80}{\pm 10}}$
& $575\,\mathord{\scriptscriptstyle\textcolor{black!60}{\pm 28}}$ 
& $112\,\mathord{\scriptscriptstyle\textcolor{black!60}{\pm 25}}$ \\

\midrule
\textbf{Column Mean} 
& $1937\,\mathord{\scriptscriptstyle\textcolor{black!60}{\pm 172}}$ 
& $\textbf{1379}\,\mathord{\scriptscriptstyle\textcolor{black!60}{\pm 120}}$ 
& $2511\,\mathord{\scriptscriptstyle\textcolor{black!60}{\pm 182}}$ 
& $600\,\mathord{\scriptscriptstyle\textcolor{black!60}{\pm 76}}$ 
& $2367\,\mathord{\scriptscriptstyle\textcolor{black!60}{\pm 185}}$ 
& $617\,\mathord{\scriptscriptstyle\textcolor{black!60}{\pm 90}}$ 
& $2001\,\mathord{\scriptscriptstyle\textcolor{black!60}{\pm 190}}$ 
& $1232\,\mathord{\scriptscriptstyle\textcolor{black!60}{\pm 84}}$ \\
\bottomrule
\end{tabular}
\caption{\textbf{Comparison study results.} 
Rows correspond to the grasp strategy variants used, and columns correspond to the interaction policy action space variants used. 
grasps and trajectories metrics are reported as mean ± standard deviation over 10 runs.}

\label{tab:experimental_box_results_marginals}
\end{table*}

%% file: tex_files/5_results_and_discussion.tex

\section{RESULTS AND DISCUSSION}
\label{sec:results_and_discussion}

\textbf{\textit{Qualitative results and analysis:}}
Figure~\ref{fig:results} illustrates the archive generated by QDTraj, where the 30 highest-performing trajectories are displayed in each image, clearly highlighting the spatial diversity captured by the method.
We observe that successful trajectories occupy different grid cells across space, indicating that QDTraj successfully explored different regions and that solutions are not clustered in a single area, as could occur with gradient-based methods. Besides, interestingly, during opening tasks, variations are concentrated on different parts of the handle, whereas during closing, diversity increases even more since the grasp is not limited to the handle region. As a result, solutions are highly diverse: we observe some non-intuitive trajectories that would likely not be discovered by a human demonstrator or classical gradient-based approaches. The algorithm’s intrinsic curiosity, enabled by MAP-Elites archive, promotes a richer diversity than human teleoperation demonstrations would usually provide without intentional effort to improve diversity. Some generated trajectories are remarkably original, highlighting the exploratory capabilities of our method, for example, closing a door without using the handle. 

We deployed successful trajectories on a real robot in a scenario illustrated in~\ref{fig:results}, where the handle is not directly reachable by the robot. This experiment highlights the benefits of having a diverse archive generated in simulation: our pipeline selects a reachable trajectory that begins by grasping the door above the handle. Although this trajectory may seem counterintuitive from a human perspective, it successfully transfers to the real robot. This scenario validates and supports our approach of generating diversity-focused trajectories.

Secondly, we can observe that trajectories differ for the same articulation, depending on the initial and final joint values of the task primitive, demonstrating that not all trajectories are reversible. Modeling the tasks with both start and end joint values is therefore essential, and it validates our physically grounded activation task classification choice. Indeed, in the context of open-ended environments, our method allows the robot to acquire expert skills for each situation, providing a foundation for robust task execution.

Lastly, and quite intuitively, a broader examination of the trajectories, considering both the experimental box and the PartNet-Mobility output archive, shows that adaptation is particularly efficient in hinge trajectory generation. All of this is generated automatically without priors as a plug-and-play method, which is a major strength of the method: it produces such a physically grounded diversity compliantly. For Partnet Faucets, the further the grasp is from the rotation axis, the larger the radius of the rotation circle, reflecting the QDTraj ability to adapt to the object’s geometry in a principled manner. The successful deployment QDTraj across different objects ( Figure \ref{fig:results}) demonstrates its generalization capabilities and further validates our approach.

\textbf{\textit{Quantitative results and analysis:}}
Table~\ref{tab:experimental_box_results_marginals} presents the quantitative analysis of the output archives obtained after 175 generations for tasks $\Gamma_1$ and $\Gamma_2$. Each cell reports the average metric value $\pm$ the standard deviation over ten independent runs, measured at the end of the learning process for all considered algorithm variants.

First, we focus on trajectory success, highlighted in green in the results table. Marginal results (row means) show that MAP-Elites-Explore consistently yields the highest number of successful trajectories, regardless of the chosen interaction policy action space. More precisely, MAP-Elites-Explore generates 5 times more successful trajectories than MAP-Elites-Success for the hinge task and 6 times more for the slider task. This indicates that cloning and evolving successful individuals across multiple parallel environments significantly accelerates exploration compared to other selection strategies.

Moreover, our adaptive policy action space produces 2.3 times more successful trajectories on average, regardless of the grasp selection strategy (column mean). More specifically, when considering the best grasp selection strategy with MAP-Elites-Explore, the adaptive action space increases the number of discovered trajectories by 3.5 times for hinge joints, highlighting the benefits of our adaptive interaction policy over state-of-the-art interaction action spaces. For slider joints, the advantage is less pronounced, with an improvement of 1.6 times. This difference can be explained by the relative simplicity of slider primitives, which can be effectively handled by simple parameterized motion primitives. In contrast, hinge tasks greatly benefit from the adaptability of our method, which enables richer and more diverse trajectories that better align with rotational constraints. Our combination contribution QD-Traj-MAP-Elites-Explore-Adaptive significantly outperforms all other configurations, increasing the maximum number of successful trajectories from 27 when using none of our proposed contributions to 2,944 when combining both contributions for hinge joints, and from 532 to 3,988 successful trajectories for slider joints.

Second, we examine the correlation between the number of discovered grasps (blue on the table~\ref{tab:experimental_box_results_marginals}) and the number of trajectories (in green). These quantities are strongly correlated: in general, more grasps lead to more trajectories. However, discovering many grasps does not necessarily guarantee a proportionally large number of trajectories, as trajectory diversity depends on both the optimization process and the trajectory generation method. Extending this analysis to the PartNet-Mobility archive we generated, the number of discovered trajectories and grasps varies across the ten objects considered (Figure~\ref{fig:APGen_traj_evolution}). On average, 704 trajectories were found per object after 50 generations (1,006 for hinge joints and 445 for slider joints). The average number of successful grasps per object is 645. These results indicate that the method generalizes well across different object types, being particularly effective for hinge joints while also performing strongly on slider joints. Variations between objects reflect differences in physical complexity and motion constraints.

Lastly, Figure~\ref{fig:APGen_traj_evolution} illustrates the evolution of archive filling with successful trajectories over generations. The $\Gamma_1$ and $\Gamma_0$ archives are plotted using a fixed grasp selection strategy, MAP-Elites-Explore, and the three varying trajectory strategies compared in this paper. With our adaptive trajectory action space, faster performance is achieved compared to all baseline methods, starting from the very first generation. This behavior validates the Adaptive action space presented in this contribution.  Interestingly, the speed of successful trajectory exploration is particularly high at the beginning of training before smoothly slowing. This behavior validates our approach and suggests the feasibility of an online Quality-Diversity setting, provided sufficient parallelization capabilities, given the high execution speed of the algorithm.

\textbf{\textit{Limits:}} A limitation of our method is that it requires an accurate physical model and inertia parameters, which is the case for our experimental object but not always for PartNet-Mobility or reconstructed scans. Additionally, the simulator occasionally produces false contacts that create spurious clusters. Although such cases are rare, they can be further minimized by adjusting certain simulator parameters (spatial discretization of the geometry, integrator time step, and stiffness and friction in the contacts).

%% file: tex_files/6_conclusions.tex

\section{CONCLUSIONS}
\input{tex_files/figures/APGen_traj_evolution}

In this work, we presented QDTraj, a method for generating diverse low-level trajectory primitives to manipulate articulated objects. By leveraging Quality-Diversity algorithms, QDTraj explores multiple ways to accomplish the same task, producing physically-grounded adaptive trajectories that can be selected in real-world conditions. Our experiments demonstrate that QDTraj generates more diverse and robust trajectory primitives than baseline methods, both on our experimental object and across 30 PartNet-Mobility articulations.

An interesting extension would be to couple QDTraj with a high-level task sequencer that orchestrates the low-level primitives we propose. Such a hierarchical, semantic approach would enable the robot to operate fully autonomously. Integrating QDTraj into this type of framework represents a promising direction for future research.

%% file: tex_files/figures/APGen_traj_evolution.tex
\begin{figure}[t]
    \centering
    \begin{minipage}{0.65\columnwidth}
        \centering
        \includegraphics[width=\linewidth]{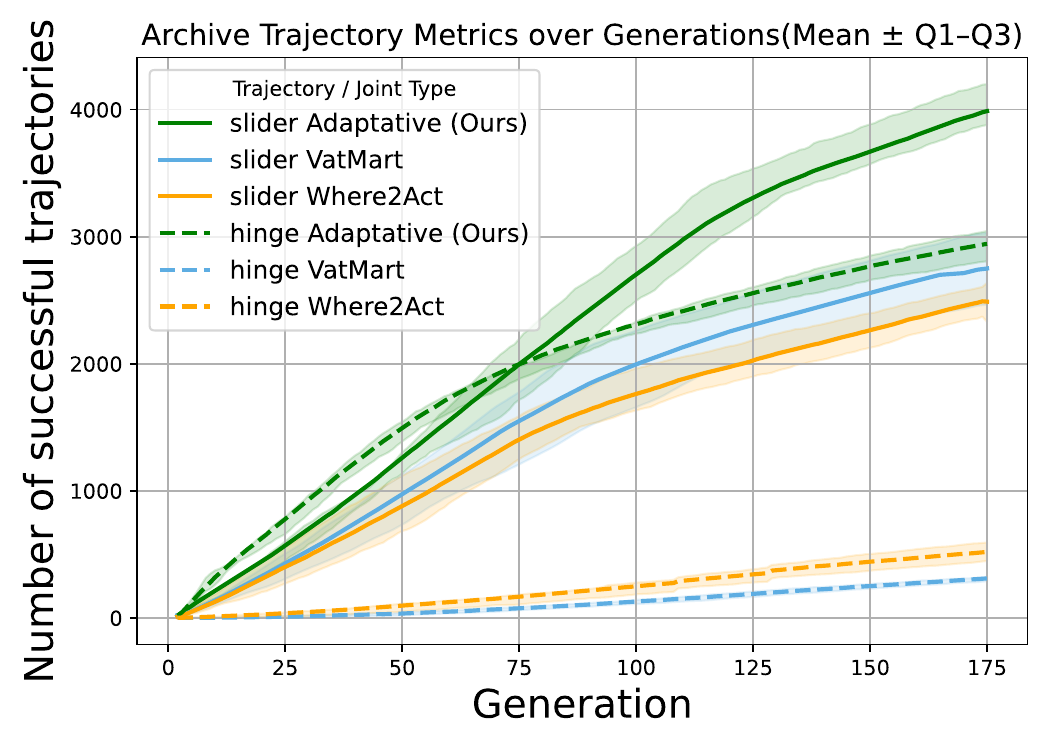}
    \end{minipage}
    \hfill
    \begin{minipage}{0.32\columnwidth}
        \centering
        \includegraphics[width=\linewidth]{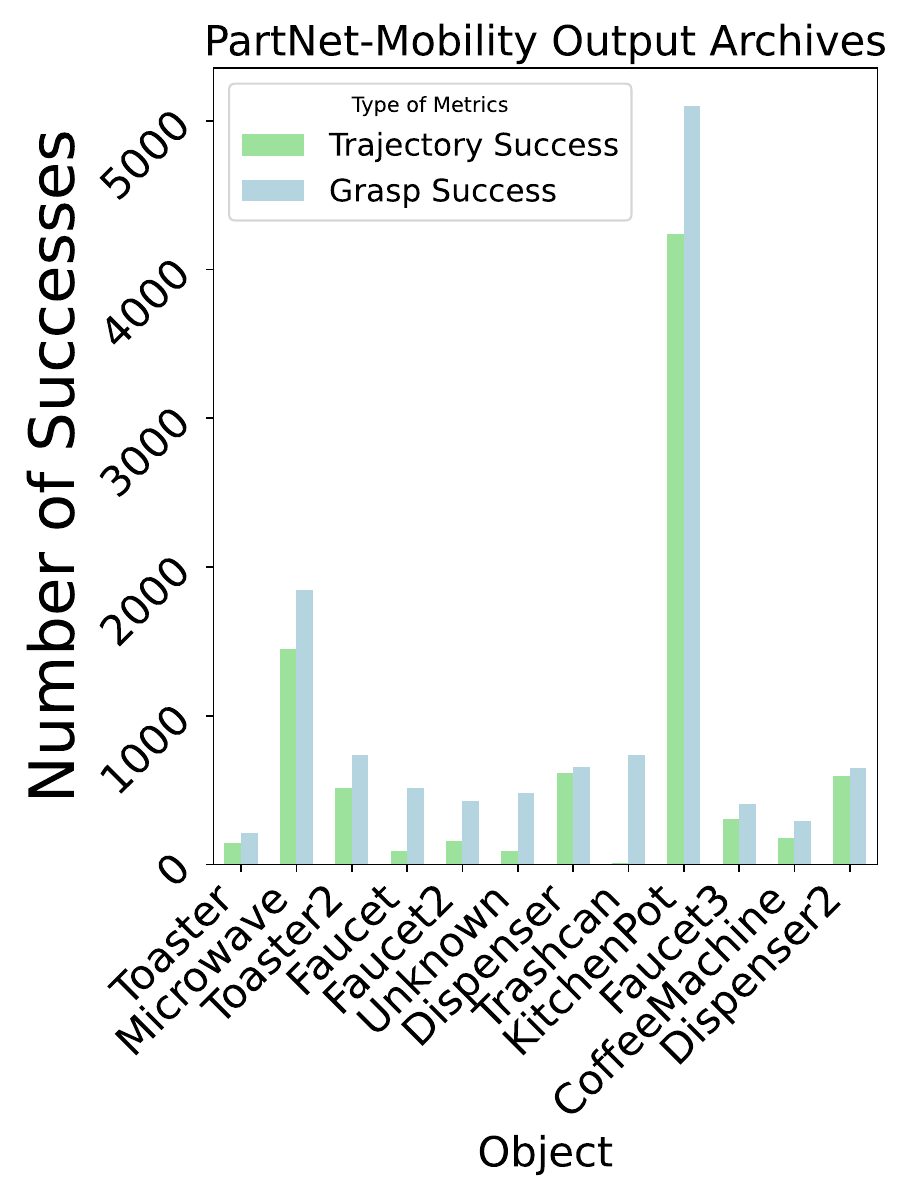}
    \end{minipage}
    \caption{\textbf{Quantitative results of QDtraj.} \textbf{Left.} Evolution of the number of successful trajectories in the archive over 175 generations for two Experimental Box primitives (averaged over 10 seeds). \textbf{Right.} Metrics of the output archive runs on 10 objects of the PartNet-Mobility Dataset.}
    \label{fig:APGen_traj_evolution}
\end{figure}

%% file: tex_files/acknowledgment.tex

\section*{ACKNOWLEDGMENT}
This work was granted access to the HPC resources of
IDRIS under the allocation 2026-AD011017431 made by
GENCI. It was also supported by the French Agence Nationale de la Recherche (Learn2Grasp, ANR-21-FAI1-0004; Tactile, ANR-25-CE33-2325), PostGenAI@Paris (ANR-23-IACL-0007, France 2030), the German Ministry
of Education and Research (BMBF; grant 01IS21080), and the European Union’s Horizon Europe Framework Programme (PILLAR, grant agreement No. 101070381; euROBIN, grant agreement No. 101070596).